\documentclass[11pt]{amsart}

\usepackage{hyperref}
\usepackage{url}
\usepackage{lipsum}
\usepackage{amsfonts}
\usepackage{graphicx}
\usepackage{epstopdf}
\usepackage{algorithmic}
\usepackage{algorithm}
\usepackage{amsmath}
\usepackage{bm}
\usepackage[margin=2.5cm]{geometry}

\mathchardef\mhyphen="2D
\newtheorem{theorem}{Theorem}
\newtheorem{proposition}{Proposition}

\newcommand{\Sec}[1]{{Section~\ref{sec:#1}}} 
\newcommand{\App}[1]{{Appendix~\ref{sec:#1}}} 
\newcommand{\Eqn}[1]{{(\ref{eq:#1})}} 
\newcommand{\Fig}[1]{{Figure~\ref{fig:#1}}} 
\newcommand{\Thm}[1]{{Theorem~\ref{thm:#1}}} 
\newcommand{\Prop}[1]{{Proposition~\ref{prop:#1}}} 
\newcommand{\Alg}[1]{{Algorithm~\ref{alg:#1}}} 
\newcommand{\As}[1]{{Assumption~{\rm\ref*{as:#1}}}} 

\newtheorem{assumption}{Assumption}[section]

\newcommand{\Real}{\mathbb{R}}

\newcommand{\Tra}{^{\sf T}} 
\newcommand{\Inv}{^{-1}} 
\def\vec{\mathop{\rm vec}\nolimits}
\newcommand{\V}[1]{{\bm{\mathbf{\MakeLowercase{#1}}}}} 
\newcommand{\VE}[2]{\MakeLowercase{#1}_{#2}} 
\newcommand{\Vtilde}[1]{{\bm{\tilde \mathbf{\MakeLowercase{#1}}}}} 

\newcommand{\M}[1]{{\bm{\mathbf{\MakeUppercase{#1}}}}} 
\newcommand{\ME}[2]{\MakeLowercase{#1}_{#2}} 

\newcommand{\Mtilde}[1]{{\bm{\tilde \mathbf{\MakeUppercase{#1}}}}} 
\newcommand{\Mn}[2]{\M{#1}^{(#2)}} 
\newcommand{\amp}{\mathop{\:\:\,}\nolimits}

\newcommand{\Kron}{\otimes} 

\title{Co-manifold learning with missing data}

\begin{document}

\author[]{Gal Mishne}
\address[Gal Mishne]{Applied Mathematics Program, Yale University, New Haven, CT, USA}
\email{gal.mishne@yale.edu}

\author[]{Eric C. Chi}
\address[Eric C. Chi]{Department of Statistics, North Carolina State University, Raleigh, NC, USA}
\email{eric\_chi@ncsu.edu}

\author[]{Ronald R. Coifman}
\address[Ronald R. Coifman]{Department of Mathematics, Yale University, New Haven, CT, USA}
\email{coifman.ronald@yale.edu}

\maketitle

\begin{abstract}
 Representation learning is typically applied to only one mode of a data matrix, either its rows or columns.
 Yet in many applications, there is an underlying geometry to both the rows and the columns.  
 We propose utilizing this coupled structure to perform co-manifold learning: uncovering the underlying geometry of both the rows and the columns of a given matrix, where we focus on a missing data setting. Our unsupervised approach consists of three components.
 We first solve a family of optimization problems to estimate a complete matrix at multiple scales of smoothness. We then use this collection of smooth matrix estimates to compute pairwise distances on the rows and columns based on a new multi-scale metric that implicitly introduces a coupling between the rows and the columns.  
  Finally, we construct row and column representations from these multi-scale metrics.  We demonstrate that our approach outperforms competing methods in both data visualization and clustering.  
\end{abstract}

\section{Introduction}
Dimension reduction plays a key role in exploratory data analysis, data visualization, clustering and classification. Techniques range from the classical PCA and nonlinear manifold learning to deep autoencoders~\cite{Tenenbaum:2000,Roweis:2000,Belkin2003,Coifman2006,Vincent2008,Rifai2011,Kingma2014}.
These techniques focus on only one mode of the data, often the observations (columns) which are are measurements in a high-dimensional feature space (rows), and exploit correlations among the features to reduce the dimension of the features and obtain the underlying low-dimensional geometry of the observations.
Yet for many data matrices, for example in gene expression studies, recommendation systems, and word-document analysis, correlations exist among both observations and features. In these cases, we seek a method that can exploit the correlations among both the rows and columns of a data matrix to better learn lower-dimensional representations of both. Biclustering methods, which extract \emph{distinct} biclusters along both rows and columns, give a partial solution to performing simultaneous dimension reduction on the rows and columns of a data matrix. Such methods, however, can break up a smooth geometry into artificial clusters. 
A more general viewpoint to consider is that data matrices possess geometric relationships between their rows (features) and columns (observations) such that both modes lie on low-dimensional \emph{manifolds}.
Furthermore, the relationships between the rows may be informed by the relationships between the columns, and vice versa. 
Several recent papers~\cite{Gavish2012,Ankenman2014,Mishne2016, Shahid2016,Mishne2017,Yair2017} exploit this coupled relationship to co-organize matrices and infer underlying row and column embeddings.

Further complicating the story is that such matrices may suffer from missing values, due to measurement corruptions and limitations.
These missing values can sabotage efforts to learn the low dimensional manifold underlying the data. 
Specifically, kernel-based methods rely on calculating a similarity matrix between observations, whose eigendecomposition yields a new embedding of the data.
As the number of missing entries grows, the distances between points are increasingly distorted, resulting in poor representation of the data in the low-dimensional space~\cite{gilbert2018unrolling}.
Matrix completion algorithms assume the data is low-rank and fill in the missing values by fitting a global linear subspace to the data. Yet, this may fail when the data lies on a nonlinear manifold.

Manifold learning in the missing data scenario has been addressed by a few recent papers.
Non-linear Principle Component Analysis (NLPCA)~\cite{scholz2005non} uses an autoencoder neural network, where the middle layer serves to learn a low-dimensional embedding of the data, and the trained autoencoder is used to fill in missing values.
Missing Data Recovery through Unsupervised Regression~\cite{carreira2011manifold} first fills in the missing data with linear matrix completion methods, then calculates a non-linear embedding of the data and incorporates this embedding in an optimization problem to fill in the missing values.
Recently \cite{gilbert2018unrolling} proposed MR-MISSING which first calculates an initial distance matrix using only non-missing entries and then uses the increase-only-metric-repair (IOMR) method to fix the distance matrix so that it is a metric from which they calculate an embedding.
None of these methods consider the co-manifold setting, where the coupled structure of the rows and the columns can be used to fill in the data, and to calculate an embedding. 

In this paper, we introduce a new method for performing joint dimension reduction on the rows and columns of a data matrix, which we term co-manifold learning, in the missing data setting. 
We build on two recent lines of work on co-organizing the rows and columns of a data matrix~\cite{Gavish2012,Mishne2016, Mishne2017} and convex optimization methods for performing co-clustering~\cite{Chi2017a,Chi2018}. 
The former provide a flexible framework for jointly organizing rows and columns but lacks algorithmic convergence guarantees. 
The latter provides convergence guarantees but does not take full advantage of the multiple scales of the data revealed in the solution.
\begin{figure}[t]
\centering
	\includegraphics[width=0.9\linewidth]{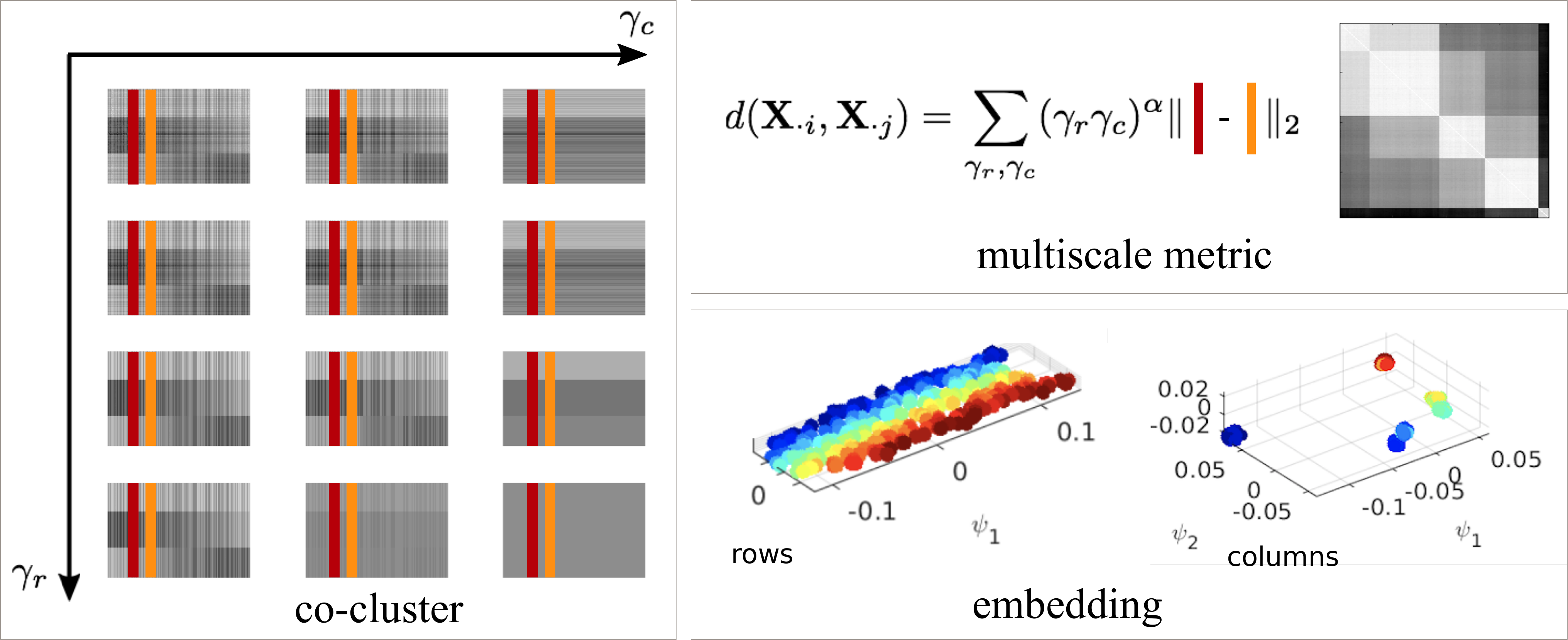}
	\caption{The three components of our approach: 1) smooth estimates of a matrix with missing entries at multiple scales via co-clustering, 2) a multi-scale metric using the smooth estimates across all scales, yielding an affinity kernel between rows/columns, and 3) nonlinear embeddings of the rows and columns. The multiscale metric between two columns (red and orange) is a weighted Euclidean distance between those columns at multiple scales, given by solving the co-clustering for increasing values of the cost parameters $\gamma_r$ and $\gamma_c$.}
	\label{fig:workflow}
\end{figure}
 Instead of inferring biclusters at a single scale, we use a multi-scale optimization framework to fill in the data, imposing smoothness on both the rows and the columns at fine to coarse scales. 
 The scale of the solution is encoded in a pair of joint cost parameters along the rows and columns. 
 The solutions to the optimization for each such pair yields a smooth estimate of the data along both the rows and columns, whose values are used to fill in the missing values of the given matrix. 
We define a new multi-scale metric based on the filled-in matrix across all scales, which we use to calculate nonlinear embeddings of the rows and columns. 
Thus our approach yields three results: a collection of smoothed estimates of the matrix, pairwise distances on the rows and columns that better estimate the geometry of the complete data matrix, and corresponding nonlinear embeddings (see \Fig{workflow}).
We will demonstrate in experimental results that our method reveals meaningful representations in coupled datasets with missing entries, whereas other methods are capable of revealing a meaningful representation only along one of the modes.

The paper is organized as follows. We present the optimization framework in \Sec{coco}, the new multi-scale metric for co-manifold learning in \Sec{manifold} and experimental results in \Sec{numerical_experiments}.

\section{Co-clustering an Incomplete Data Matrix}
\label{sec:coco}

We seek a collection of complete matrix approximations of a partially observed data matrix $\M{X} \in \Real^{m \times n}$ that have been smoothed along their row and columns to varying degrees. This collection will serve in computing row and column multi-scale metrics to better estimate the row and column pairwise distances of the complete data matrix. Let $[m]$ denote the set of indices $\{1, \ldots, m\}$, and let $\Theta \subseteq [m] \times [n]$ be a subset of the indices that correspond to observed entries of $\M{X}$, and let $\mathcal{P}_\Theta$ denote the projection operator of $m \times n$ matrices onto an index set $\Theta$, i.e.\@ $[P_\Theta(\M{X})]_{ij}$ is $\ME{X}{ij}$ if $(i,j) \in \Theta$ and is 0 otherwise.

We seek a minimizer $\M{U}(\gamma_r, \gamma_c)$ of the following function.
\begin{eqnarray}
\label{eq:obj}
f(\M{U}; \gamma_r, \gamma_c) & = & \frac{1}{2}\lVert \mathcal{P}_\Theta(\M{X}) - \mathcal{P}_\Theta(\M{U}) \rVert_{\text{F}}^2 + \gamma_rJ_r(\M{U}) + \gamma_cJ_c(\M{U}).
\end{eqnarray}
The quadratic term quantifies how well $\M{U}$ approximates $\M{X}$ on the observed entries, while the two roughness penalties, $J_r(\M{U})$ and $J_c(\M{U})$, incentivize smoothness across the rows and columns of $\M{U}$. The nonnegative parameters $\gamma_r$ and $\gamma_c$ tune the tradeoff between how well $\M{U}$ agrees with $\M{X}$ over $\Theta$ and how smooth $\M{U}$ is with respect to its rows and columns. By tuning $\gamma_r$ and $\gamma_c$, we obtain  estimates of $\M{X}$ at varying scales of row and column smoothness. 

In this paper, we use roughness penalties of the following forms
\begin{eqnarray*}
J_r(\M{U}) & = & \sum_{(i,j) \in \mathcal{E}_r} \Omega(\lVert \M{U}_{i \cdot} -  \M{U}_{j \cdot} \rVert_2) \quad\text{and}\quad
J_c(\M{U}) = \sum_{(i,j) \in \mathcal{E}_c} \Omega(\lVert \M{U}_{\cdot i} -  \M{U}_{\cdot j} \rVert_2),
\end{eqnarray*}
where $\M{U}_{i \cdot}$ $(\M{U}_{\cdot i})$ denotes the $i$th row (column) of the matrix $\M{U}$.
The index sets $\mathcal{E}_r$ and $\mathcal{E}_c$ denote the edge sets of row and column graphs that encode a preliminary data-driven assessment of the similarities between rows and columns of the data matrix. The function $\Omega$, which maps $[0, \infty)$ into $[0, \infty)$, will be explained shortly. The convergence properties of our co-clustering procedure will rely on the following two assumptions.

\begin{assumption}
\label{as:connectedness}
The row and column graphs $\mathcal{E}_r$ and $\mathcal{E}_c$ are connected, i.e.\@ the row graph is connected if for any pair of rows, indexed by $i$ and $j$ with $i \neq j$, there exists a sequence of indices $i \rightarrow k \rightarrow \cdots \rightarrow l \rightarrow j$ such that $(i,k), \ldots, (l,j) \in \mathcal{E}_r$. A column graph is connected under analogous conditions.
\end{assumption}

\begin{assumption}
\label{as:omega}
The function $\Omega : [0, \infty) \mapsto [0, \infty)$
 is (i) concave and continuously differentiable on $(0, \infty)$, (ii) vanishes at the origin, i.e. $\Omega(0) = 0$, (iii) is increasing on $[0, \infty)$, and (iv) has finite directional derivative at the origin. 
\end{assumption}

Variations on the optimization problem of minimizing \Eqn{obj} have been previously proposed in the literature. When there is no data missing, i.e.\@ $\Theta = [m] \times [n]$ and $\Omega$ is a linear mapping, minimizing the objective in \Eqn{obj} produces a convex biclustering problem \cite{Chi2017a}. Additionally, if either $\gamma_r$ or $\gamma_c$ is zero, then we obtain convex clustering \cite{PelDeSuy2005, HocVerBac2011, LinOhlLju2011, Chi2015}. If we take $\Omega$ to be a nonlinear concave function, problem \Eqn{obj} reduces to an instance of concave penalized regression-based clustering \cite{PanShenLiu2013, Marchetti2014, Wu2016}.

Replacing $J_r(\M{U})$ and $J_c(\M{U})$ by quadratic row and column Laplacian penalties 
\begin{eqnarray*}
J_r(\M{U}) & = & \frac{1}{2}\sum_{(i,j) \in \mathcal{E}_r} \lVert \M{U}_{i \cdot} -  \M{U}_{j \cdot} \rVert^2_2 \quad\text{and}\quad
J_c(\M{U}) = \frac{1}{2}\sum_{(i,j) \in \mathcal{E}_c} \lVert \M{U}_{\cdot i} -  \M{U}_{\cdot j} \rVert^2_2,
\end{eqnarray*}
gives a version of matrix completion on graphs \cite{Kalofolias2014, Rao2015}. \cite{Shahid2016} also use row and column Laplacian penalties to perform joint linear dimension reduction on the rows and columns of the data matrix. Our work generalizes both \cite{Shahid2016} and \cite{Chi2017a} in that we seek the flexibility of performing non-linear dimension reduction on the rows and columns of the data matrix and seek more general manifold organization than co-clustered structure.

\subsection{Co-Clustering Algorithm}
\label{sec:alg}

\begin{algorithm}[t]
\begin{algorithmic}[1]
  \caption{\textsc{co-cluster-missing}($\mathcal{P}_\Theta(\M{X}), \gamma_r,\gamma_c$)}
  \label{alg:MM}
\STATE Initialize $\M{U}_0, \tilde{w}_{r, ij},$ and $\tilde{w}_{c, ij}$
\REPEAT
\STATE $\Mtilde{X} \gets \mathcal{P}_{\Theta}(\M{X}) + \mathcal{P}_{\Theta^c}(\M{U}_{t})$
\STATE $\left \{\M{U}_{t+1},n_r,n_c \right\}\gets \textsc{convex-bicluster}\left(\Mtilde{X}, \gamma_r, \gamma_c, \{\tilde{w}_{r, ij}\}, \{\tilde{w}_{c, ij}\}\right)$
\STATE $\tilde{w}_{r, ij} \gets \Omega'(\lVert \M{U}_{t+1,i \cdot} - \M{U}_{t+1,j \cdot}\rVert_2)$ for all $(i,j) \in \mathcal{E}_r$
\STATE $\tilde{w}_{c, ij} \gets \Omega'(\lVert \M{U}_{t+1,\cdot i} - \M{U}_{t+1,\cdot j}\rVert_2)$ for all $(i,j) \in \mathcal{E}_c$
\UNTIL{convergence}
\STATE Return $\left\{\M{U}(\gamma_r,\gamma_c)  = \M{U}_t, \Mtilde{X}, n_r, n_c\right\}$
\end{algorithmic}
\end{algorithm}
We now introduce a majorization-minimization (MM) algorithm \cite{Sun2017} for solving the minimization in \Eqn{obj}. 
The basic strategy behind an MM algorithm is to convert a hard optimization problem into a sequence of simpler ones. The MM principle requires majorizing the objective function $f(\M{U})$ by a surrogate function $g(\M{u} \mid \Mtilde{u})$ anchored at $\Mtilde{u}$.  Majorization is a combination of the tangency condition $g(\M{u} \mid \Mtilde{u}) =  f( \Mtilde{u})$ and the domination condition $g(\M{U} \mid \Mtilde{u})  \geq f(\M{U})$ for all $\M{U} \in \Real^{m \times n}$.  The associated MM algorithm is defined by the iterates $\M{u}_{t+1} = \underset{\M{U}}{\arg \min}\; g(\M{U} \mid \M{U}_{t})$. It is straightforward to verify that the MM iterates generate a descent algorithm driving the objective function downhill, i.e.\@ that $f(\M{u}_{t+1}) \leq f(\M{U}_{t})$ for all $t$. 

The following function
\begin{eqnarray*}
g(\M{U} \mid \Mtilde{U})
& = & \frac{1}{2}\lVert \Mtilde{X} - \M{U} \rVert_{\text{F}}^2 + \gamma_r\sum_{(i,j) \in \mathcal{E}_r} \tilde{w}_{r,ij} \lVert \M{U}_{i \cdot} - \M{U}_{j \cdot }\rVert_2 + \gamma_c\sum_{(i,j) \in \mathcal{E}_c} \tilde{w}_{c,ij} \lVert \M{U}_{\cdot i} - \M{U}_{\cdot j}\rVert_2 + \kappa \nonumber
\end{eqnarray*}
majorizes our objective function \Eqn{obj} at $\Mtilde{U}$, where $\kappa$ is a constant that does not depend on $\M{U}$ and
$\tilde{w}_{r,ij}$ and $\tilde{w}_{c,ij}$ are weights that depend on $\Mtilde{U}$, i.e.
\begin{eqnarray}
\label{eq:weights}
\tilde{w}_{r,ij} & = & \Omega'(\lVert \Mtilde{U}_{i \cdot} - \Mtilde{U}_{j \cdot}\rVert_2) \quad\text{and}\quad
\tilde{w}_{c,ij} \amp = \amp \Omega'(\lVert \Mtilde{U}_{\cdot i} - \Mtilde{U}_{\cdot j}\rVert_2).
\end{eqnarray}
We give a detailed derivation of this majorization in \App{majorization}.

Minimizing $g(\M{U} \mid \Mtilde{U})$ is equivalent to minimizing the objective function of the convex biclustering problem for which efficient algorithms have been introduced \cite{Chi2017a}. Thus, in the $t+1$th iteration, our MM algorithm
solves a convex biclustering problem where the missing values in $\M{X}$ have been replaced with the values of $\Mtilde{U} = \M{U}_{t}$ and the weights $\tilde{w}_{r,ij}$ and $\tilde{w}_{c,ij}$ have been computed based on $\Mtilde{U} = \M{U}_{t}$ according to \Eqn{weights}.

\Alg{MM} summarizes our MM algorithm, \textsc{co-cluster-missing}, which returns a smooth output matrix $\M{U}(\gamma_r,\gamma_c)$, a filled-in matrix $\Mtilde{X} = \mathcal{P}_\Theta(\M{X}) + \mathcal{P}_{\Theta^c}(\M{U}(\gamma_r, \gamma_c))$ as well as $n_r$ and $n_c$, which are respectively the number of distinct rows and distinct columns in $\M{U}(\gamma_r, \gamma_c)$. The \textsc{co-cluster-missing} algorithm has the following convergence guarantee whose proof is in \App{convergence}.
\begin{proposition}
\label{prop:convergence}
Under \As{connectedness} and \As{omega}, the sequence $\M{U}_{t}$ generated by \Alg{MM} has at least one limit point, and all limit points are stationary points of \Eqn{obj}.
\end{proposition}

In the rest of this paper, we use the following function $\Omega$ which satisfies \As{omega}
\begin{eqnarray*}
\label{eq:snowflake}
\Omega(z) & = & \frac{1}{2}\int_0^z \frac{1}{\sqrt{\zeta} + \epsilon}d\zeta,
\end{eqnarray*}
where $\epsilon$ is a small positive number, e.g.\@ $10^{-12}$. We briefly explain the rationale in our choice. By the monotone convergence theorem, as $\epsilon$ tends to zero, $\Omega(z)$ converges to the mapping $z \mapsto \sqrt{z}$. Thus, $\Omega(\lVert \M{U}_{i \cdot} - \M{U}_{j \cdot} \rVert_2)$ approximates a snowflake metric $d(\M{U}_{i \cdot}, \M{U}_{j \cdot}) = \sqrt{\lVert \M{U}_{i \cdot} - \M{U}_{j \cdot} \rVert_2}$. When the approximate snowflake metric is employed in the penalty term, small differences between rows and columns are penalized significantly more than larger differences resulting in more aggressive smoothing of small noisy variations and less smoothing of more significant systematic variations. Note that the weights are continuously updated throughout the optimization as opposed to the fixed weights in~\cite{Chi2017a}. This introduces a notion of the scale of the solution into the weights.

\begin{algorithm}[t]
\begin{algorithmic}[1]
  \caption{Co-manifold learning on an Incomplete Data Matrix}
  \label{alg:CM}
\STATE Initialize $\mathcal{E}_r, \mathcal{E}_c$
\STATE Set $d(\M{X}_{\cdot i },\M{X}_{\cdot j} )=0$ and $d(\M{X}_{\cdot i },\M{X}_{\cdot j} )=0$
\STATE Set $n_r = m, n_c=n, k = k_0$, and  $l = l_0$
\WHILE{ $n_r > 1$}
\WHILE{ $n_c > 1$}
\STATE $\left \{\Mn{U}{l,k}, \Mtilde{X}^{(l,k)}, n_r,n_c \right\} \gets$  \textsc{co-cluster-missing}$\left(\mathcal{P}_\Theta(\M{X}), \gamma_r=2^{l} ,\gamma_c=2^{k}\right)$
\STATE Update row distances:  $d\left(\M{X}_{ i \cdot},\M{X}_{ j \cdot}\right)  \mathrel{{+}{=}} d\left(\Mtilde{X}^{(l,k)}_{ i \cdot},\Mtilde{X}^{(l,k)}_{ j \cdot}\right)$
\STATE Update column distances: $d\left(\M{X}_{\cdot i },\M{X}_{\cdot j}\right)  \mathrel{{+}{=}} d\left(\Mtilde{X}^{(l,k)}_{\cdot i },\Mtilde{X}^{(l,k)}_{\cdot j}\right)$
\STATE $k \gets k + 1$
 \ENDWHILE
\STATE  $l \gets l + 1$
 \ENDWHILE
  \STATE Calculate affinities $\M{A}_r(\M{X}_{ i \cdot},\M{X}_{ j \cdot})$ and $\M{A}_c(\M{X}_{ \cdot i },\M{X}_{ \cdot j })$ 
   \STATE Calculate embeddings $\Psi_r, \Psi_c$
 \end{algorithmic}
\end{algorithm}

\subsection{Co-clustering at multiple scales}
Initializing \Alg{MM} is very important as the objective function in \Eqn{obj} is not convex.
The matrix $\Mn{U}{0}$ is initialized to be the mean of all non-missing values.
The connectivity graphs $\mathcal{E}_r$ and $\mathcal{E}_c$ are initialized at the beginning using $k$-nearest-neighbor graphs, and remain fixed throughout all considered scales. If we observed the complete matrix, employing a sparse Gaussian kernel is a natural way to
quantify the local similarity between pairs of rows and pairs of columns.
The challenge is that we do not have the complete data matrix $\M{X}$ but
only the partially observed one $\mathcal{P}_{\Theta}(\M{X})$.
Therefore, we rely only on the observed values to calculate the k-nearest-neighbor graph, using the distance used by~\cite{ram2013image} in an image inpainting problem.

To obtain a collection of estimates at multiple scales, we need to solve the optimization problem for pairs of $\gamma_r,\gamma_c$.
We start with small values of $\gamma_r=2^{l_0}$ and $\gamma_c=2^{k_0}$, where $l_0,k_0<0$.
We calculate the co-clustering (\Alg{MM}) and obtain the smooth estimate $\Mn{U}{l_0,k_0} = \M{U}(2^{l_0}, 2^{k_0})$, the filled-in data matrix $\Mtilde{X}^{(l_0,k_0)}$, and $n_r$ and $n_c$ which are the number of distinct row and column clusters, respectively, identified at that scale. 
Keeping $\gamma_r$ fixed, we keep increasing $\gamma_c$ by power of 2 and biclustering the data until the algorithm converges to one cluster along the columns. 
We then increase $\gamma_r$ by power of 2 and reset $\gamma_c=2^{k_0}$.
We repeat this procedure at increasing scales of $\gamma_r=2^{l},\quad \gamma_c=2^{k}$, until $n_r=n_c=1$, indicating we have converged to a single global bicluster. 
Thus we traverse a solution surface at logarithmic scale~\cite{Chi2018b}. 
This yields a collection of filled-in matrices at all scales $\left\{\Mtilde{X}^{(l,k)}\right\}_{l,k}$.

\section{Co-manifold learning}
\label{sec:manifold}
Kernel-based manifold learning relies on constructing a ``good" similarity measure between points, and a dimension reduction method based on this similarity. 
The eigenvectors of these kernels is typically used as the new low-dimensional coordinates for the data.
Here we leverage having calculated an estimate of the filled-in matrix at multiple scales $\left\{\Mtilde{X}^{(l,k)}\right\}_{l,k}$, to define a new metric between rows and columns. 
This metric will encompass all bi-scales as defined by joint pairs of optimization cost parameters $\gamma_r,\gamma_c$. 
Given a new metric we employ diffusion maps to obtain a new embedding of the rows and columns. Note that other methods can be used for embedding based on our new metric. The full algorithm is given in \Alg{CM}.

\subsection{Multi-scale metric}
\label{sec:metric}
We define a new metric to estimate the geometry both locally and globally of the complete data matrix. For a given pair $\gamma_r,\gamma_c$, we calculate the Euclidean distance between rows for the filled-in matrix at that joint scale, weighted by the cost parameters:
\begin{eqnarray*}
\label{eq:emdPartitionInpaint}
d\left(\Mtilde{X}^{(l,k)}_{i \cdot},\Mtilde{X}^{(l,k)}_{j \cdot}\right) & = & (\gamma_r \gamma_c)^\alpha \Vert \Mtilde{X}^{(l,k)}_{i \cdot}-\Mtilde{X}^{(l,k)}_{j \cdot} \Vert_2
\end{eqnarray*}
where $\Mtilde{X}^{(l,k)} = \mathcal{P}_{\Theta}(\M{X}) + \mathcal{P}_{\Theta^c}(\M{U}^{(l,k)})$. 
Having solved for multiple paris from the solution surface, we sum over all the distances to obtain a multi-scale distance on the data rows:
\begin{eqnarray*}
\label{eq:emdPartitionInpaintSum}
d(\M{X}_{i \cdot },\M{X}_{j \cdot }) & = & \sum_{l,k} d\left(\Mtilde{X}^{(l,k)}_{i \cdot },\Mtilde{X}^{(l,k)}_{j \cdot }\right).
\end{eqnarray*}
An analogous multi-scale distance is computed for pairs of columns.

This metric takes advantage of solving the optimization for multiple pairs of cost parameters and filling in the missing values with increasingly smooth estimates. 
Note that if there are no missing values, this metric is just the Euclidean pairwise distance scaled by a scalar, so that we recover the embedding of the complete matrix. 
In our simulations, we set $\alpha=-1/2$ to favor local over global structure. 
As opposed to the partition-tree based metric of~\cite{Mishne2017}, this metric takes into account all joint scales of the data as the matrix $\M{U}$ is smoothed across rows and columns \emph{simultaneously}, thus fully taking advantage of the coupling between both modes.

\subsection{Diffusion maps}
Having calculated a multi-scale metric on the rows and columns throughout the joint optimization procedure, we can now construct a pair of low-dimensional embeddings based on these distances. Specifically we use diffusion maps~\cite{Coifman2006}, but any dimension reduction technique relying on the construction of a distance kernel could be used instead. 
We briefly review the construction of the diffusion maps for the rows (features) of a matrix but the same can be applied to the columns (observations). 
Given a distance between two rows of the matrix $d(\M{X}_{i \cdot } , \M{X}_{j \cdot } )$, we construct an affinity kernel on the rows. 
We choose a Gaussian kernel, but other kernels can be considered depending on the application:
\begin{eqnarray*}
 \M{A}[i, j] & = & \exp\{ - d^2(\M{X}_{i \cdot } , \M{X}_{j \cdot } )/ \sigma^2 \},
\end{eqnarray*}
where $\sigma$ is a scale parameter.
This kernel enhances locality, as pairs of samples whose distance exceed $\sigma$ have negligible affinity. 
One possible choice for $\sigma$ is to be the median of pairwise distances within the data.

We derive a row-stochastic matrix by normalizing the rows of $\M{A}$:
\begin{eqnarray*}
\M{P} & = & \M{D}\Inv\M{A}, 
\end{eqnarray*}
where $\M{D}$ is a diagonal matrix whose elements are given by $\M{D}[i,i] = \sum_j \M{A}[i,j]$.
The eigendecomposition of $\M{P}$ yields a sequence of decreasing eigenvalues: $1 = \lambda_0\geq\lambda_1\geq ...$, and right eigenvectors $\{\psi_\ell\}_\ell$.
Retaining only the first $d$ eigenvalues and eigenvectors, the mapping $\Psi$ embeds the rows into the Euclidean space $\mathbb{R}^{d}$:
\begin{eqnarray*}
\label{eq:diffusion_map}
\Psi: \M{X}_{i \cdot } \rightarrow \big( \lambda_1\psi_1(i), \lambda_2\psi_2(i),..., \lambda_{d}\psi_{d}(i)\big)\Tra.
\end{eqnarray*}
The embedding integrates the local connections found in the data into a global representation, which enables visualization of the data, organizes the data into meaningful clusters, and identifies outliers and singular samples.
This embedding is also equipped with a noise-robust distance, the diffusion distance.
For more details on diffusion maps, see~\cite{Coifman2006}.

\section{Numerical Experiments}
\label{sec:numerical_experiments}

We applied our approach to three datasets, and evaluated results both qualitatively and quantitatively:
\begin{itemize}
	\item {\bf{linkage}} A synthetic dataset with a one-dimensional manifold along the rows and a two-dimensional manifold along the columns. Let $\{z_i\}_{i=1}^{N_1}\in\mathbb{R}^3$ be points along a helix and let $\{y_j\}_{j=1}^{N_2}\in\mathbb{R}^3$ be a two dimensional surface. We analyze the matrix of Euclidean distances between the two spatially distant sets of points to reveal the underlying geometry of both rows and columns, 
	\begin{equation*}
	\M{X}[i,j] = \Vert z_i - y_j \Vert_2.
	\end{equation*}
	Other functions of the distance can also be used such as the elastic or Coulomb potential operator~\cite{coifman2011harmonic}. Missing values correspond to having access to only some of the distances between pairs of points across the two sets. Note that this is unlike MDS as we do not have pairwise distances between all datapoints, but rather distances between two sets of points with different geometries. 
	\item {\bf{linkage2}} A synthetic dataset with a clustered structure along the rows and a two-dimensional manifold along the columns. Let $\{x_i\}_{i=1}^{N_1}\in\mathbb{R}^3$ be composed of points in 3 Gaussian clouds in 3D and let $\{y_j\}_{j=1}^{N_2}\in\mathbb{R}^3$ be a two dimensional surface as before. 
	\item {\bf{lung500}} A real-world dataset composed of 56 lung cancer patients and their gene expression~\cite{lee2010biclustering}. We selected the 500 genes with the greatest variance from the original collection of 12,625 genes. Subjects belong to one of four subgroups; they are either normal subjects (Normal) or have been diagnosed with one of three types of cancers: pulmonary carcinoid tumors (Carcinoid), colon metastases (Colon), and small cell carcinoma (Small Cell).
\end{itemize}

\begin{figure}[t]
\centering
	\includegraphics[width=1.0\linewidth]{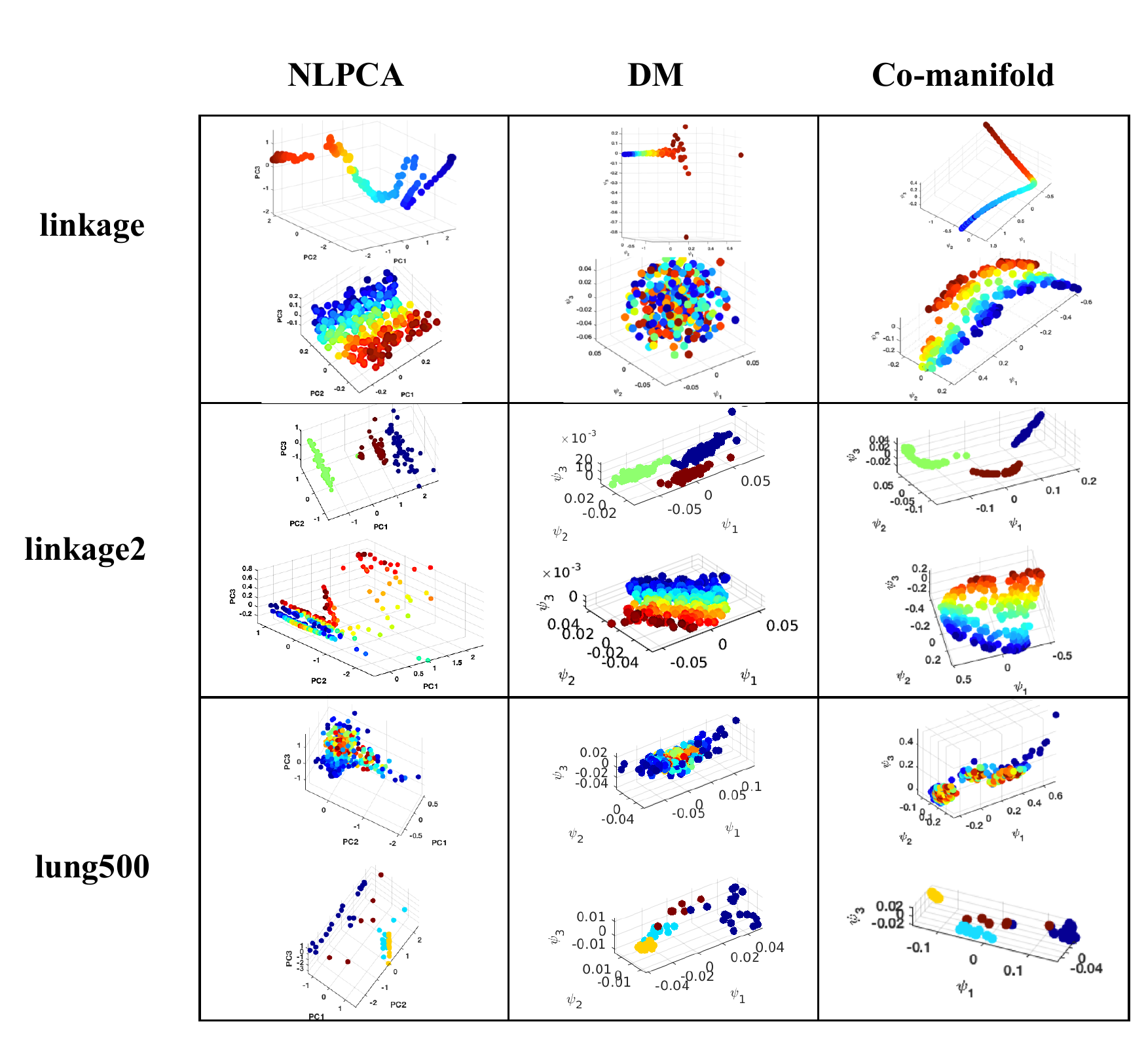}
	\caption{Comparing row and column embeddings of NLPCA, DM, Co-manifold, for three datasets with 50\% missing entries. For each dataset, top / bottom plot is embedding of rows / columns of $\M{X}$.}
	\label{fig:results}
\end{figure}

The rows and columns of the data matrix are randomly permuted so their natural order does not play a role in inferring the geometry.
In \Fig{results}, we compare our embeddings to both NLPCA with missing data completion~\cite{scholz2005non} and Diffusion maps (DM)~\cite{Coifman2006} on the missing data, where both methods are applied to each mode separately, while our co-manifold approach takes into account the coupled geometry.
Comparing to Diffusion maps demonstrates how missing values corrupt the embedding.
In all examples 50\% of the entries are missing.
For each of the three methods we display the embedding for both the rows (top) and the columns (bottom),
Both NLPCA and DM reveal the underlying 2D surface structure on the rows in only one of the linkage datasets, and err greatly on the other. DM correctly infers a 1D path for the  {\bf{linkage}} dataset but it is increasingly noisy. For NLPCA the 1D embedding is not as smooth and clean as the embedding inferred by the co-manifold approach.
Our method reveals the 2D surface in both cases.
For the {\bf{lung500}} data, NLPCA and DM embed the cancer samples such that the normal subjects (yellow) are close to the Colon type (cyan), whereas our method separates the normal subjects from the cancer types. This is due to taking into account the coupled structure of the genes and the samples.
All three methods reveal a smooth manifold structure to the genes, which is different than the assumed clustered structure a biclustering method would infer. 
For plots presenting the datasets and filled-in values at multiple scales see \App{exp}.

\begin{figure}[t]
\centering
	\includegraphics[width=0.95\linewidth]{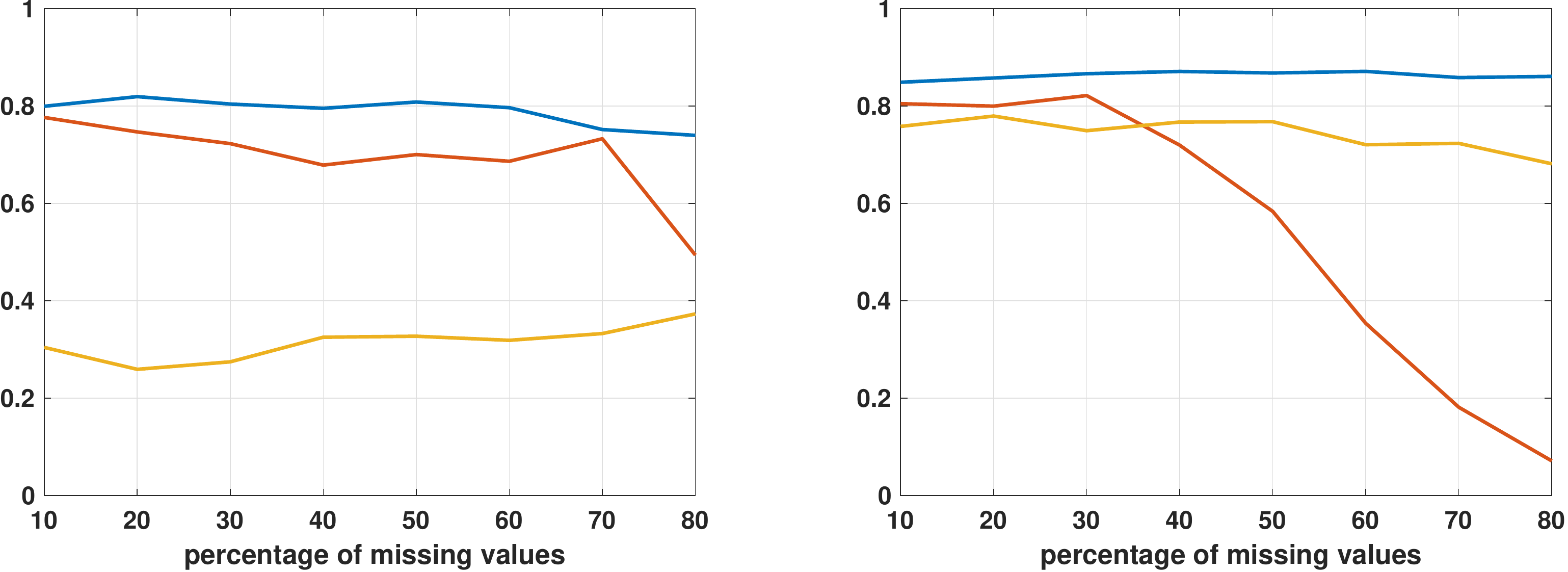}
	\caption{Comparing k-means clustering applied to embedding of data using ours (blue), diffusion maps of missing data matrix (red), and NLPCA (yellow) for increasing percentages of missing values. We calculate the adjusted Rand Index compared to the ground-truth labels of (left) the 4 cancer types for the {\textbf{lung500}} dataset, and (right) 3 Gaussian clusters of the {\textbf{linkage2}} dataset}
	\label{fig:ARI}
\end{figure}
Manifold learning is not used only for data visualization but also for calculating new data representations that can then be used for signal processing and machine learning tasks. The left panel of 
\Fig{ARI} compares clustering the embedding of the cancer patients in {\bf{lung500}} by each method for increasing percentage of missing values in the data, where we averaged over 30 realizations of missing entries.
We use the Adjusted Rand Index (ARI)~\cite{Hubert1985}, to measure the similarity between the k-means clustering of the embedding and the ground-truth labels.
Our embedding (blue plot) gives the best clustering result and its performance is only slightly degraded by increasing the percentage of missing values, as opposed to Diffusion maps (red plot). This demonstrates that the metric we calculate is a good estimate of the metric of the complete data matrix. NLPCA (yellow plot) performs worst.

The right panel of \Fig{ARI} compares clustering the embedding of the three Gaussian clusters in {\bf{linkage2}} for increasing percentage of missing values in the data, where we averaged over 30 realizations of missing entries.
Our embedding (blue plot) gives the best clustering result and its performance is unaffected by increasing the percentage of missing values, as opposed to Diffusion maps (red plot) which is greatly degraded by the missing values. NLPCA (yellow plot) does not perform as well as our approach, with performance decreasing as the percentage of missing values increases.

\section{Conclusions}
\label{sec:conclusions}

In this paper we presented a new method for learning nonlinear manifold representations of both the rows and columns of a matrix with missing data. 
We proposed a new optimization problem to obtain a smooth estimate of the missing data matrix, and solved this problem for different values of the cost parameters, which encode the smoothness scale of the estimate along the rows and columns. We leverage calculating these multi-scale estimates into a new metric that aims to capture the geometry of the complete data matrix. This metric is then used in a kernel-based manifold learning technique to obtain new representations of both the rows and the columns.
In future work we will investigate additional metrics in a general co-manifold setting and relate them to optimal transport problem and Earth Mover's Distance~\cite{Leeb2013}.

\bibliography{references}
\bibliographystyle{IEEEtran}

\newpage

\appendix


\section{Derivation of Majorization}
\label{sec:majorization}

We first construct a majorization of the data-fidelity term. It is easy to verify that the following function of $\M{U}$
\begin{eqnarray}
\label{eq:mm1}
g_1(\M{U} \mid \Mtilde{U}) & = & \frac{1}{2}\lVert \Mtilde{X} - \M{U} \rVert_{\text{F}}^2,
\end{eqnarray}
where $\Mtilde{X} = \mathcal{P}_\Theta(\M{X}) + \mathcal{P}_{\Theta^c}(\Mtilde{U})$, 
majorizes the data-fidelity term $\frac{1}{2}\lVert \mathcal{P}_{\Theta}(\M{X}) - \mathcal{P}_{\Theta}(\M{U}) \rVert_{\text{F}}^2$ at $\Mtilde{U}$.

We next construct a majorization of the penalty term. Recall that the first-order Taylor approximation of a differentiable concave function provides a tight bound on the function.  
Therefore, under \As{omega}, we have the following inequality
\begin{eqnarray*}
\Omega(z) & \leq & \Omega(\tilde{z}) + \Omega'(\tilde{z})(z - \tilde{z}), \quad\quad\text{for all $z, \tilde{z} \in [0, \infty)$}.
\end{eqnarray*}
Thus, we can majorize the penalty term $\gamma_rJ_r(\M{U}) + \gamma_cJ_c(\M{U})$ with the function
\begin{eqnarray}
\label{eq:mm2}
g_2(\M{U} \mid \Mtilde{U}) & = & \gamma_r\sum_{(i,j) \in \mathcal{E}_r} \tilde{w}_{r,ij} \lVert \M{U}_{i \cdot} - \M{U}_{j \cdot }\rVert_2 + \gamma_c\sum_{(i,j) \in \mathcal{E}_c} \tilde{w}_{c,ij} \lVert \M{U}_{\cdot i} - \M{U}_{\cdot j}\rVert_2 + \kappa,
\end{eqnarray}
where $\kappa$ is a constant that does not depend on $\M{U}$ and $\tilde{w}_{r,ij}$ and $\tilde{w}_{c,ij}$~\Eqn{weights} are weights that depend on $\Mtilde{U}$.
The sum of functions \Eqn{mm1} and \Eqn{mm2}
\begin{eqnarray}
\label{eq:majorization}
g(\M{U} \mid \Mtilde{U}) & = & g_1(\M{U} \mid \Mtilde{U}) + g_2(\M{U} \mid \Mtilde{U}) \\
& = & \frac{1}{2}\lVert \Mtilde{X} - \M{U} \rVert_{\text{F}}^2 + \gamma_r\sum_{(i,j) \in \mathcal{E}_r} \tilde{w}_{r,ij} \lVert \M{U}_{i \cdot} - \M{U}_{j \cdot }\rVert_2 + \gamma_c\sum_{(i,j) \in \mathcal{E}_c} \tilde{w}_{c,ij} \lVert \M{U}_{\cdot i} - \M{U}_{\cdot j}\rVert_2 + \kappa \nonumber
\end{eqnarray}
majorizes our objective function \Eqn{obj} at $\Mtilde{U}$.

\section{Convergence}
\label{sec:convergence}

The MM algorithm generates a sequence of iterates that has at least one limit point, and
the limit points are stationary points of the objective function
\begin{eqnarray}
\label{eq:objective_function}
f(\M{U}) & = & \frac{1}{2}\lVert \mathcal{P}_{\Theta}(\M{X}) - \mathcal{P}_{\Theta}(\M{U}) \rVert_{\text{F}}^2 + \gamma_rJ_r(\M{U}) + \gamma_cJ_c(\M{U}).
\end{eqnarray}
To reduce notational clutter, we suppress the dependency of $f$ on $\gamma_r$ and $\gamma_c$ since they are fixed during \Alg{MM}.
We prove \Prop{convergence} in three stages. First, we show that all limit points of the MM algorithm are fixed points of the MM algorithm map. Second, we show that fixed points of the MM algorithm are stationary points of $f$ in \Eqn{objective_function}. Finally, we show that the MM algorithm has at least one limit point.

\subsection{Limit points are fixed points}

The convergence theory of MM algorithms relies on the properties of the algorithm map $\psi(\M{u})$ that returns the next iterate given the last iterate. For easy reference, we state a simple version of Meyer's monotone convergence theorem \cite{Meyer1976}, which is instrumental in proving convergence in our setting.
\begin{theorem}\label{thm:MM_limit_points}
  Let $f(\M{U})$ be a continuous function on a domain $S$ and
   $\psi(\M{U})$ be a continuous algorithm map from $S$ into $S$ satisfying
 $f(\psi(\M{U})) < f(\M{U})$ for all $\M{U} \in S$ with $\psi(\M{U}) \neq \M{U}$. Then all limit points of the iterate sequence $\M{U}_{k} = \psi(\M{U}_{k-1})$ are fixed points of $\psi(\M{U})$.
\end{theorem}

In order to apply \Thm{MM_limit_points}, we need to identify elements in the assumption with specific functions and sets corresponding to the problem of minimizing \Eqn{objective_function}. Throughout the following proof, it will sometimes be convenient to work with the column major vectorization of a matrix. The vector $\V{b} = \vec(\M{B})$ is obtained by stacking the columns of $\M{B}$ on top of each other.

{\bf The function $f$:} Take $\mathcal{S} = \Real^{m\times n}$ and $f: \mathcal{S} \mapsto \Real$ to be the objective function in \Eqn{objective_function} and majorize $f$ with $g(\M{U} \mid \Mtilde{U})$ given in \Eqn{majorization}. 
The function $f$ is continuous. Let $\psi(\Mtilde{u}) = \underset{\M{u}}{\arg\min}\; g(\M{u} \mid \Mtilde{u})$ denote the algorithm map for the MM algorithm. Since $g(\M{U} \mid \Mtilde{U})$ is strongly convex in $\M{U}$, it has a unique global minimizer. Consequently, $f(\psi(\M{U})) < f(\M{U})$ for all $\psi(\M{U}) \neq \M{U}$.

{\bf Continuity of the algorithm map $\psi$:}  Continuity of $\psi$ follows from the fact that the solution to the convex biclustering problem is jointly continuous in the weights and data matrix \cite{Chi2017a}[Proposition 2]. The weight $\tilde{w}_{r,ij}(\Mtilde{U}) = \Omega'(\lVert \M{U}_{i\cdot} - \M{U}_{j\cdot}\rVert_2)$ is a continuous function of $\Mtilde{U}$, since $\Omega'$ is continuous according to \As{omega}. The weight $\tilde{w}_{c,ij}(\Mtilde{U})$ is likewise continuous in $\Mtilde{U}$. The data matrix passed into the convex biclustering algorithm is $\Mtilde{X} = \mathcal{P}_{\Theta}(\M{X}) + \mathcal{P}_{\Theta^c}(\Mtilde{U})$, which is a continuous function of $\Mtilde{U}$ since the projection mapping $\mathcal{P}_{\Theta^c}$ is continuous.

\subsection{Fixed points are stationary points}

Let $\M{L}_{ij} = (\V{e}_i - \V{e}_j)\Tra\Kron\M{I}$ and $\Mtilde{L}_{ij} = \M{I} \Kron (\V{e}_i - \V{e}_j)\Tra$, where $\Kron$ denotes the Kronecker product. Then
\begin{eqnarray*}
\vec(\M{U}_{i\cdot} - \M{U}_{j \cdot}) & = & \M{L}_{ij}\V{u} \quad \quad \text{and} \quad\quad 
\vec(\M{U}_{\cdot i} - \M{U}_{\cdot j}) \amp = \amp \Mtilde{L}_{ij}\V{u}.
\end{eqnarray*}
The directional derivative of $f$ in the direction $\V{v}$ at a point $\V{u}$ is given by
\begin{eqnarray*}
\Omega'(\lVert \M{L}_{ij}\V{u} \rVert_2; \V{v}) & = & \begin{cases}
\Omega'(\lVert \M{L}_{ij}\V{u}\rVert_2) \langle \M{L}_{ij}\V{v}, \frac{\M{L}_{ij}\V{u}}{\lVert \M{L}_{ij}\V{u} \rVert_2}\rangle & \M{L}_{ij}\V{u} \neq \V{0} \\
\Omega'(\lVert \M{L}_{ij}\V{u}\rVert_2)\lVert \M{L}_{ij} \V{v} \rVert_2 & \text{otherwise.}
\end{cases}
\end{eqnarray*}

A point $\V{u}$ is a stationary point of $f$, if for all direction vectors $\V{v}$
\begin{eqnarray*}
0 & \leq & \langle \mathcal{P}_\Theta(\V{u} - \V{x}), \V{v} \rangle + \gamma_r \sum_{(i,j) \in \mathcal{E}_r} \Omega'(\lVert \M{L}_{ij}\V{u}\rVert_2; \V{v}) + \gamma_c \sum_{(i,j) \in \mathcal{E}_c} \Omega'(\lVert \Mtilde{L}_{ij}\V{u}\rVert_2; \V{v}),
\end{eqnarray*}
where $\mathcal{P}_\Theta(\V{u} - \V{x}) = \vec(\mathcal{P}_\Theta(\M{U}) - \mathcal{P}_\Theta(\M{X}))$.

A point $\V{u}$ is a fixed point of $\psi$, if $\V{0}$ is in the subdifferential of $g(\V{u} \mid \V{u})$, i.e.\begin{eqnarray}
\label{eq:fixed_point}
\V{0}\in\{\mathcal{P}_\Theta(\V{u} - \V{x})\} + \gamma_r\sum_{(i,j) \in \mathcal{E}_r} \Omega'(\lVert \M{L}_{ij}\V{u} \rVert_2) \partial \lVert \M{L}_{ij}\V{u} \rVert_2 + \gamma_c\sum_{(i,j) \in \mathcal{E}_c} \Omega'(\lVert \Mtilde{L}_{ij}\V{u} \rVert_2) \partial \lVert \Mtilde{L}_{ij}\V{u} \rVert_2,
\end{eqnarray}
where the set on the right is the subdifferential $\partial g(\V{u} \mid \V{u})$.

If $\M{L}_{ij}\V{u} \neq \V{0}$, then $\partial \lVert \M{L}_{ij}\V{u} \rVert_2 =  \left \{\M{L}_{ij}\Tra\frac{\M{L}_{ij}\V{u}}{\lVert \M{L}_{ij}\V{u} \rVert_2}\right\}$. On the other hand, if $\M{L}_{ij}\V{u} = \V{0}$, then $\partial \lVert \M{L}_{ij}\V{u} \rVert_2 = \partial \lVert \V{0} \rVert_2 = \{ \V{d} : \lVert \V{d} \rVert_2 \leq 1\}$.

Fix an arbitrary direction vector $\V{v}$. The inner product of $\V{v}$ with an element in the set on right hand side of \Eqn{fixed_point} is given by
\begin{eqnarray}
\label{eq:directional_derivative}
\langle \mathcal{P}_\Theta(\V{u} - \V{x}), \V{v} \rangle + \gamma_r\sum_{(i,j) \in \mathcal{E}_r} \Omega'(\lVert \M{L}_{ij}\V{u} \rVert_2) \langle \V{d}_{ij}, \V{v} \rangle + \gamma_c\sum_{(i,j) \in \mathcal{E}_c} \Omega'(\lVert \Mtilde{L}_{ij}\V{u} \rVert_2) \langle \V{d}_{ij}, \V{v} \rangle,
\end{eqnarray}
where $\V{d}_{ij} \in \partial \lVert \M{L}_{ij}\V{u} \rVert_2$ and $\Vtilde{d}_{ij} \in \partial \lVert \Mtilde{L}_{ij}\V{u} \rVert_2$.

Then the supremum of the right hand side of \Eqn{directional_derivative} over all $\V{d}_{ij} \in \partial \lVert \M{L}_{ij}\V{u} \rVert_2$ and $\Vtilde{d}_{ij} \in \partial \lVert \Mtilde{L}_{ij}\V{u} \rVert_2$ is nonnegative, because $\V{0} \in \partial g(\V{u} \mid \V{u})$. Consequently, all fixed points of $\psi$ are also stationary points of $f$.

\subsection{The MM iterate sequence has a limit point}

To ensure the existence of a limit point, we show that the function $f$ is coercive, i.e.\@ $f(\M{U}_t) \rightarrow \infty$ for any sequence $\lVert \M{U}_t \rVert_{\text{F}} \rightarrow \infty$. Recall that according to \As{connectedness} we assume that the row and column edge sets $\mathcal{E}_r$ and $\mathcal{E}_c$ form connected graphs. Therefore, $J_r(\M{U}) = J_c(\M{U}) = 0$ if and only if $\M{U} = a\V{1}\V{1}\Tra$ \cite[Proposition 3]{Chi2017a}.
The edge-incidence matrix of the column graph $\M{\Phi}_c \in \Real^{\lvert \mathcal{E}_c \rvert \times n}$ encodes its connectivity and is defined as
\begin{eqnarray*}
\ME{\phi}{c,li} = \begin{cases}
1 & \text{If node $i$ is the head of edge $l$,} \\
-1 & \text{If node $i$ is the tail of edge $l$,} \\
0 & \text{otherwise.}
\end{cases}
\end{eqnarray*}
The row edge-incidence matrix $\M{\Phi}_r \in \Real^{\lvert \mathcal{E}_r \rvert \times m}$ is defined similarly.
Assume that $\Theta$ non-empty, i.e.\@ at least one entry of the matrix has been observed. Finally, assume that $\Omega$ is also coercive.

Note that any sequence $\M{U}_t = a_t \V{1}\V{1}\Tra + \M{B}_t$ where $\langle \M{B}_t, \V{1}\V{1}\Tra \rangle = 0$. Note that $J_r(\M{U}_t) = J_r(\M{B}_t)$ and $J_c(\M{U}_t) = J_c(\M{B}_t)$. 
Let $\M{U}_t$ be a diverging sequence, i.e.\@ $\lVert \M{U}_t \rVert_{\text{F}} \rightarrow \infty$. There are two cases to consider. 

{\bf Case I:} Suppose that $\lVert \M{B}_t \rVert_{\text{F}} \rightarrow \infty$. Let
\begin{eqnarray*}
\M{L} & = & \begin{pmatrix}
\M{I} \Kron \M{\Phi}_r \\
\M{\Phi}_c \Kron \M{I}
\end{pmatrix} \amp \in \amp \Real^{\lvert \mathcal{E}_r \rvert m + \lvert \mathcal{E}_c \rvert n \times mn},
\end{eqnarray*}
and let $\sigma_{\min}$ denote the smallest singular value of $\M{L}$. Note that the null space of $\M{L}$ is the span of $\V{1}$. Therefore, since $\langle \V{1}, \V{b}_t \rangle = 0$

\begin{eqnarray}
\label{eq:lower_bound1}
\lVert \M{L}\V{b}_t \rVert_2 & \geq & \sigma_{\min}\lVert \M{B}_t  \rVert_{\text{F}}.
\end{eqnarray}

Also note that
\begin{eqnarray*}
\M{L}\V{b}_t & = & \begin{pmatrix}
\vec(\M{\Phi}_r\M{B}_t) \\
\vec(\M{B}_t\M{\Phi}_c\Tra)
\end{pmatrix}.
\end{eqnarray*}
Since the mapping $\V{x} = \begin{pmatrix} \V{x}_1\Tra & \V{x}_2\Tra \end{pmatrix}\Tra \mapsto \max\{ \lVert \V{x}_1 \rVert_2, \lVert \V{x}_2\rVert_2\}$ is a norm, and all finite dimensional norms are equivalent, there exists some $\eta > 0$ such that
\begin{eqnarray}
\label{eq:lower_bound2}
\eta \lVert \M{L}\V{b}_t \rVert_2 & \leq & \max\left\{\lVert \M{\Phi}_r \M{B}_t \rVert_{\text{F}}, \lVert \M{B}_t\M{\Phi}_c\Tra \rVert_{\text{F}}\right\}.
\end{eqnarray}

By the triangle inequality
\begin{eqnarray} 
\label{eq:lower_bound3}
\max\left\{
\lVert \M{\Phi}_r \M{B}_t \rVert_{\text{F}}, \lVert \M{B}_t\M{\Phi}_c\Tra \rVert_{\text{F}} \right\} & \leq & \max\left\{\sum_{(i,j) \in \mathcal{E}_r} \lVert \M{L}_{ij}\V{b}_t \rVert_2, \sum_{(i,j) \in \mathcal{E}_c} \lVert \Mtilde{L}_{ij}\V{b}_t \rVert_2\right\}.
\end{eqnarray}
Let $M = \max\{\lvert \mathcal{E}_r \rvert, \lvert \mathcal{E}_c \rvert\}$ then
\begin{eqnarray}
\label{eq:lower_bound4}
 \max\left\{\sum_{(i,j) \in \mathcal{E}_r} \lVert \M{L}_{ij}\V{b}_t \rVert_2, \sum_{(i,j) \in \mathcal{E}_c} \lVert \Mtilde{L}_{ij}\V{b}_t \rVert_2\right\}
 \leq M \max\left\{\max_{(i,j) \in \mathcal{E}_r} \lVert \M{L}_{ij} \V{b}_t \rVert_2, \max_{(i,j) \in \mathcal{E}_c} \lVert \Mtilde{L}_{ij} \V{b}_t \rVert_2\right\}.
\end{eqnarray}

Putting inequalities \Eqn{lower_bound1}, \Eqn{lower_bound2}, \Eqn{lower_bound3}, and \Eqn{lower_bound4} together gives us
\begin{eqnarray}
\frac{\eta\sigma_{\min}}{M} \lVert \M{B}_t \rVert_{\text{F}} & \leq & \max \left\{
\max_{(i,j) \in \mathcal{E}_r} \lVert \M{L}_{ij}\V{b}_t \rVert_2, \max_{(i,j) \in \mathcal{E}_c} \lVert \Mtilde{L}_{ij} \V{b}_t \rVert_2 \right\}.
\end{eqnarray}
Since $\Omega$ is increasing according to \As{omega}, it follows that
\begin{eqnarray}
\label{eq:lower_bound5}
\Omega \left ( \frac{\eta\sigma_{\min}}{M} \lVert \M{B}_t \rVert_{\text{F}}\right) & \leq & 
\max \left\{\Omega\left(
\max_{(i,j) \in \mathcal{E}_r} \lVert \M{L}_{ij}\V{b}_t \rVert_2\right), \Omega\left(\max_{(i,j) \in \mathcal{E}_c} \lVert \Mtilde{L}_{ij} \V{b}_t \rVert_2 \right)\right\}.
\end{eqnarray}
Inequality \Eqn{lower_bound5} implies that
\begin{eqnarray*}
\min\{\gamma_r, \gamma_c\}M \Omega \left ( \frac{\eta\sigma_{\min}}{M} \lVert \M{B}_t \rVert_{\text{F}}\right) & \leq &
\min\{\gamma_r, \gamma_c\}\max\left\{J_r(\M{U}_t), J_c(\M{U}_t) \right\} \\
& \leq &
\gamma_rJ_r(\M{U}_t) + \gamma_cJ_c(\M{U}_t).
\end{eqnarray*}
Consequently, since $\Omega$ is increasing and $\lVert \M{B}_t \rVert_{\text{F}} \rightarrow \infty$ implies that $f(\M{U}_t) \rightarrow \infty$.

{\bf Case II:} Suppose $\lVert \M{B}_t \rVert_{\text{F}} \leq B$ for some $B$. Then $\lvert a_t \rvert \rightarrow \infty$.
Note that we have the following inequality
\begin{eqnarray*}
f(\M{U}_t) & \geq & \sum_{(i,j) \in \Theta} (\ME{X}{ij} - \VE{b}{k,ij} - a_t)^2 \\
& \geq & \sum_{(i,j) \in \Theta} a_t^2 - 2a_t(\ME{X}{ij} - \VE{b}{k,ij}) \\
& = & \lvert \Theta \rvert a_t^2 - 2a_t \sum_{(i,j) \in \Theta} (\ME{X}{ij} - \VE{b}{k,ij}) \\
& \geq & \lvert \Theta \rvert a_t^2 - 2a_t \underset{\lVert \M{B}_t \rVert_{\text{F}} \leq B}{\sup}\sum_{(i,j) \in \Theta} (\ME{X}{ij} - \VE{b}{k,ij}) \\
& = & \lvert \Theta \rvert \left [ a_t^2 - 2a_t C\right] \\
& = & \lvert \Theta \rvert \left [ (a_t - C)^2 - C^2 \right],
\end{eqnarray*}
where $C = \lvert \Theta \rvert\Inv\underset{\lVert \M{B}_t \rVert_{\text{F}} \leq B}{\sup}\sum_{(i,j) \in \Theta} (\ME{X}{ij} - \VE{b}{k,ij})$.

The function $(a_t - C)^2$ diverges since $\lvert a_t \rvert \rightarrow \infty$. Therefore, the function $f$ is coercive.

\section{Filling in missing data}
\label{sec:exp}
\begin{figure}[t]
\centering
	\includegraphics[width=0.5\linewidth]{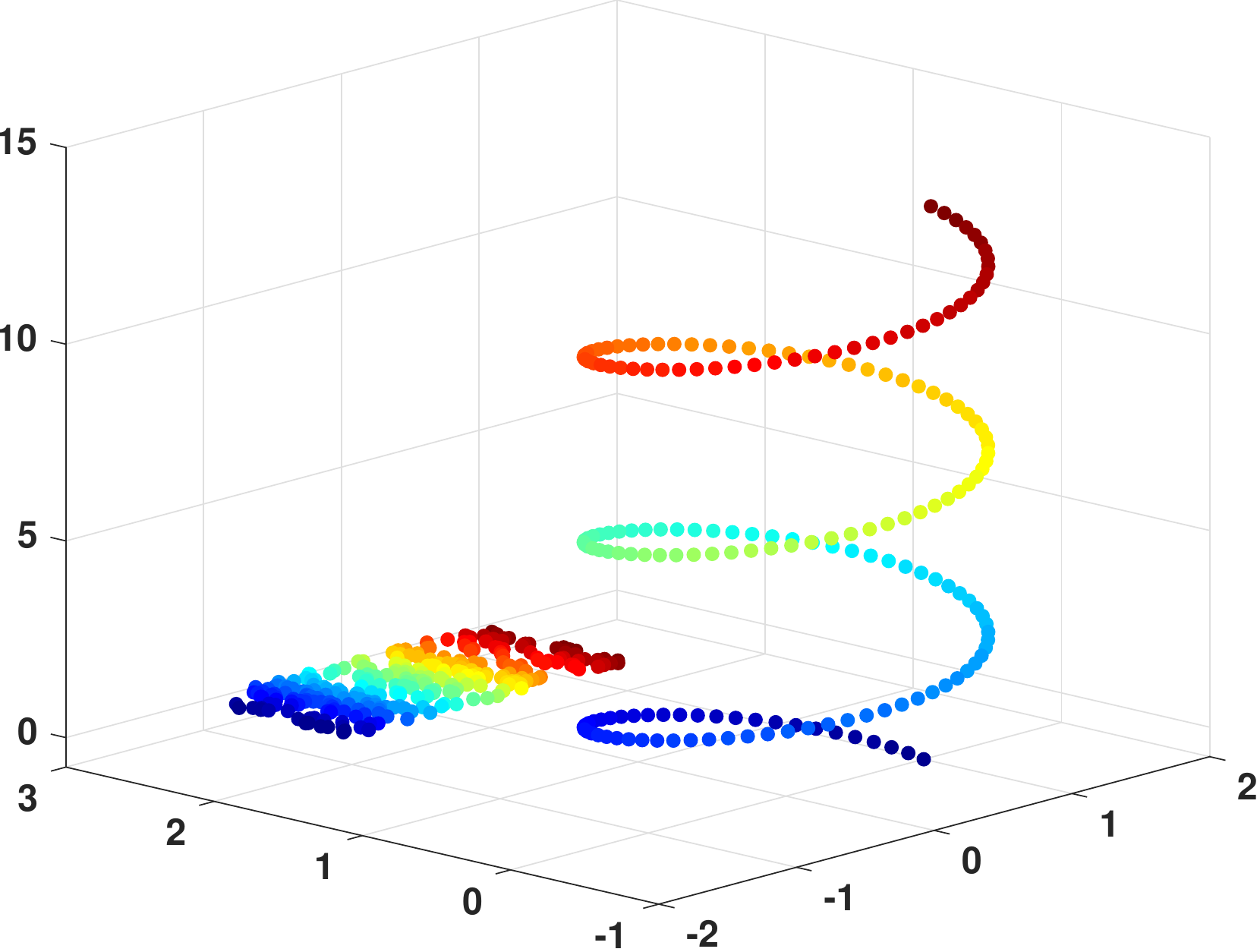}
	\caption{Points in 3D used to generate the Euclidean distance matrix $\M{X}$ in the {\textbf{linkage}} dataset. Rows correspond to the helix, columns to the 2D surface. The embedding of rows and columns in \Fig{results} are colored corresponding to the points here.}
	\label{fig:pot}
\end{figure}

\begin{figure}[t]
\centering
	\includegraphics[width=0.95\linewidth]{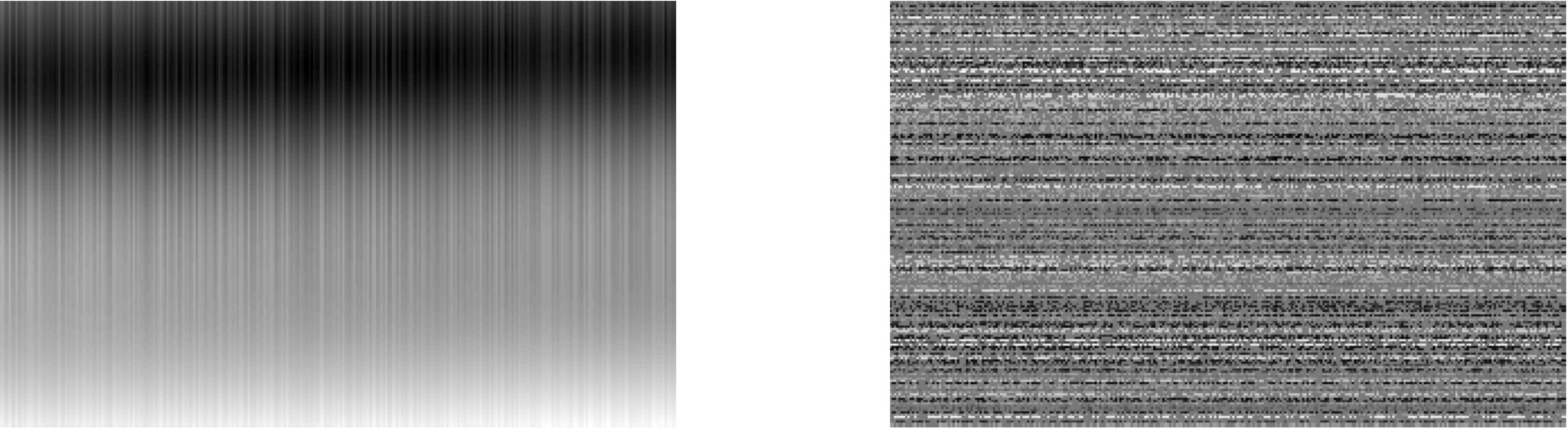}
	\caption{ {\textbf{linkage}} dataset: (Left) Complete matrix $\M{X}$. (Right) Matrix whose rows and columns and columns have been permuted and $50\%$ of the values have been removed.}
	\label{fig:pot_mix}
\end{figure}

\begin{figure}[t]
\centering
	\includegraphics[width=0.95\linewidth]{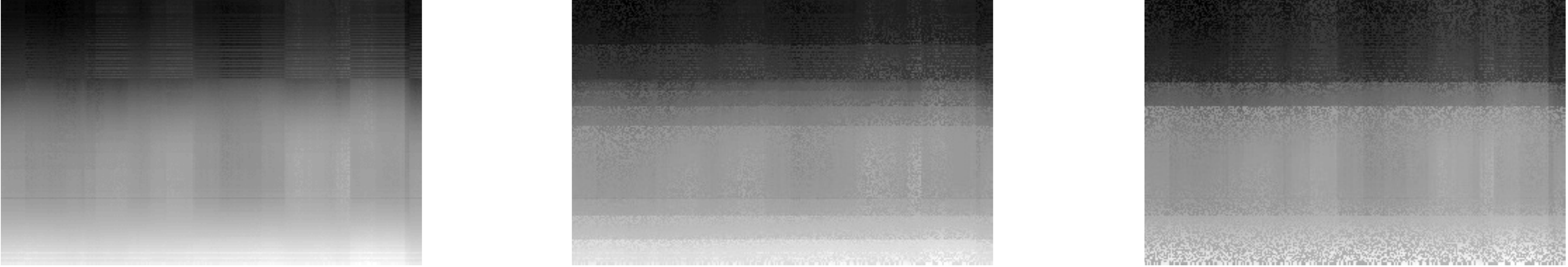}
	\caption{ {\textbf{linkage}} dataset: Filled-in matrices $\Mtilde{X}$ at multiple scales: $\Mtilde{X}^{(-3,-2)}$,$\Mtilde{X}^{(1,0)}$,$\Mtilde{X}^{(5,2)}$. Rows and columns have been reordered based on the manifold embedding following~\cite{Ankenman2014}.}
	\label{fig:pot_fill}
\end{figure}

\begin{figure}[t]
\centering
	\includegraphics[width=0.5\linewidth]{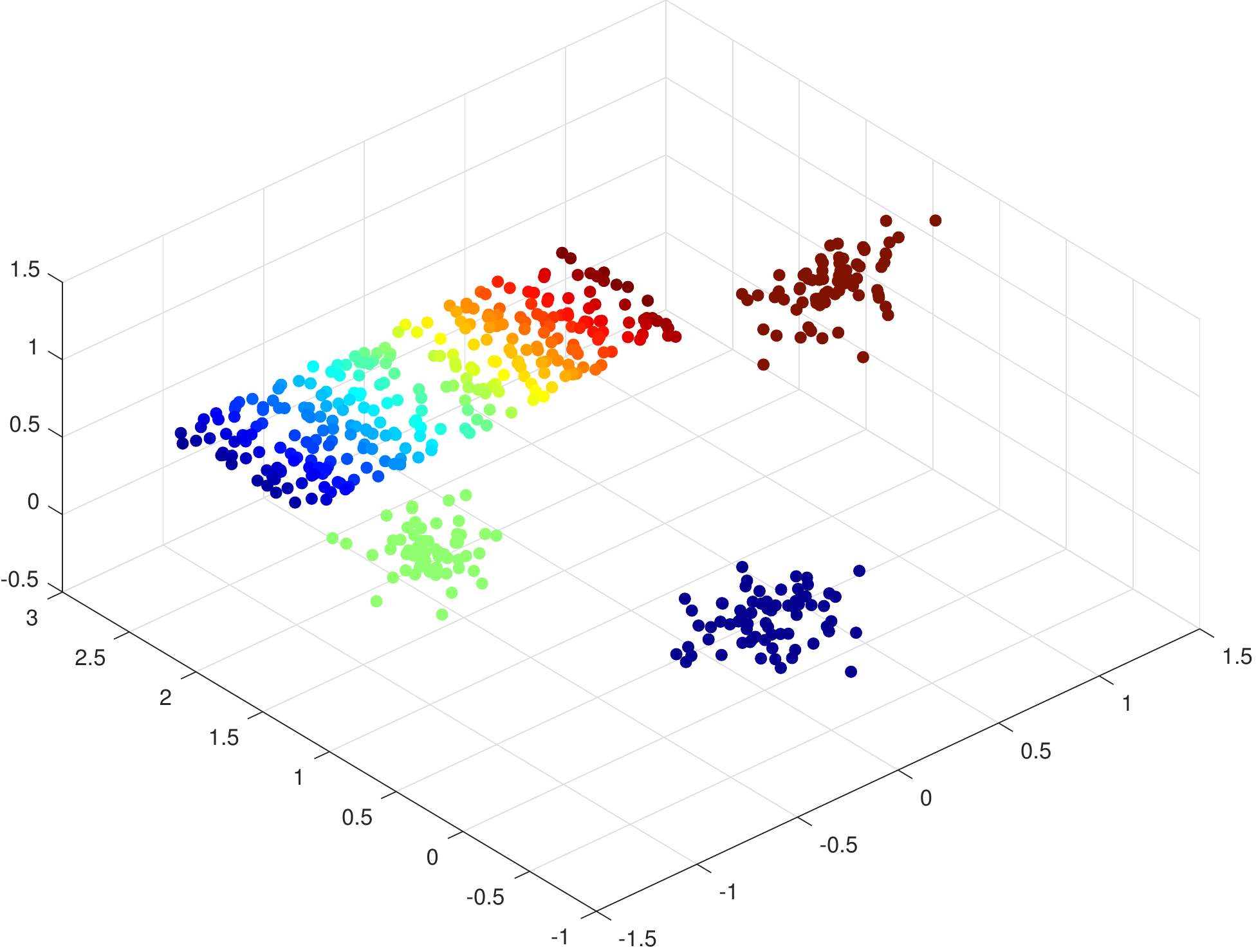}
	\caption{Points in 3D used to generate the Euclidean distance $\M{X}$ in the {\textbf{linkage2}} dataset. Rows correspond to the three 3D Gaussians, columns to the 2D surface. The embedding of rows and columns in \Fig{results} are colored corresponding to the points here.}
	\label{fig:pot2}
\end{figure}

\begin{figure}[t]
\centering
	\includegraphics[width=0.8\linewidth]{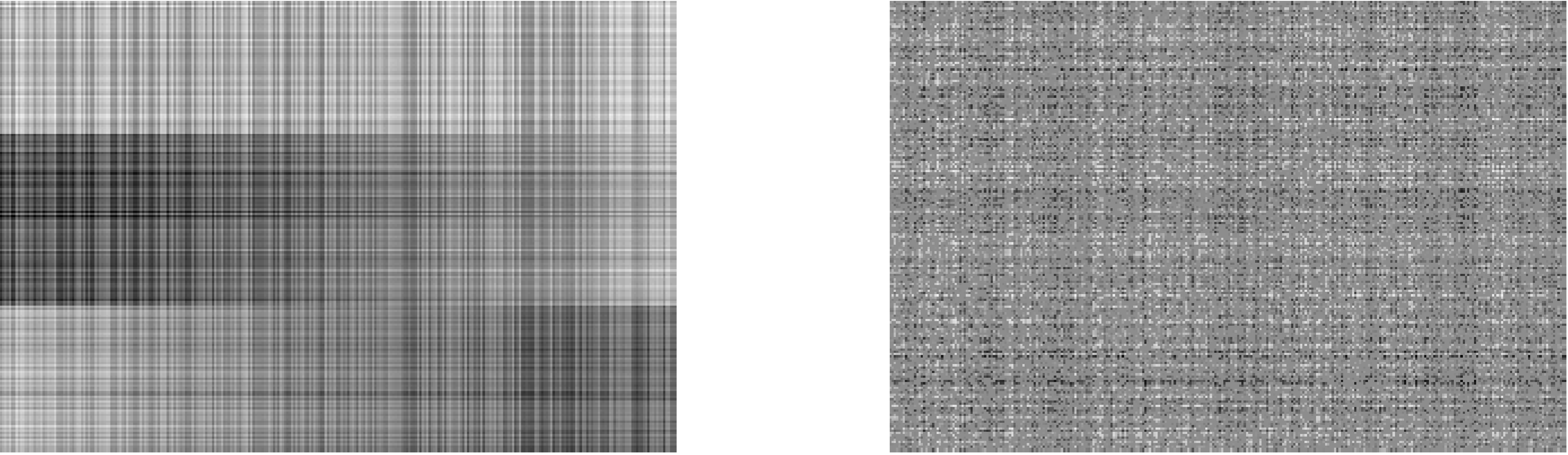}
	\caption{{\textbf{linkage2}} dataset: (Left) Complete matrix $\M{X}$. (Right) Matrix whose rows and columns and columns have been permuted and $50\%$ of the values have been removed.}
	\label{fig:pot2_mix}
\end{figure}

\begin{figure}[t]
\centering
	\includegraphics[width=0.35\linewidth]{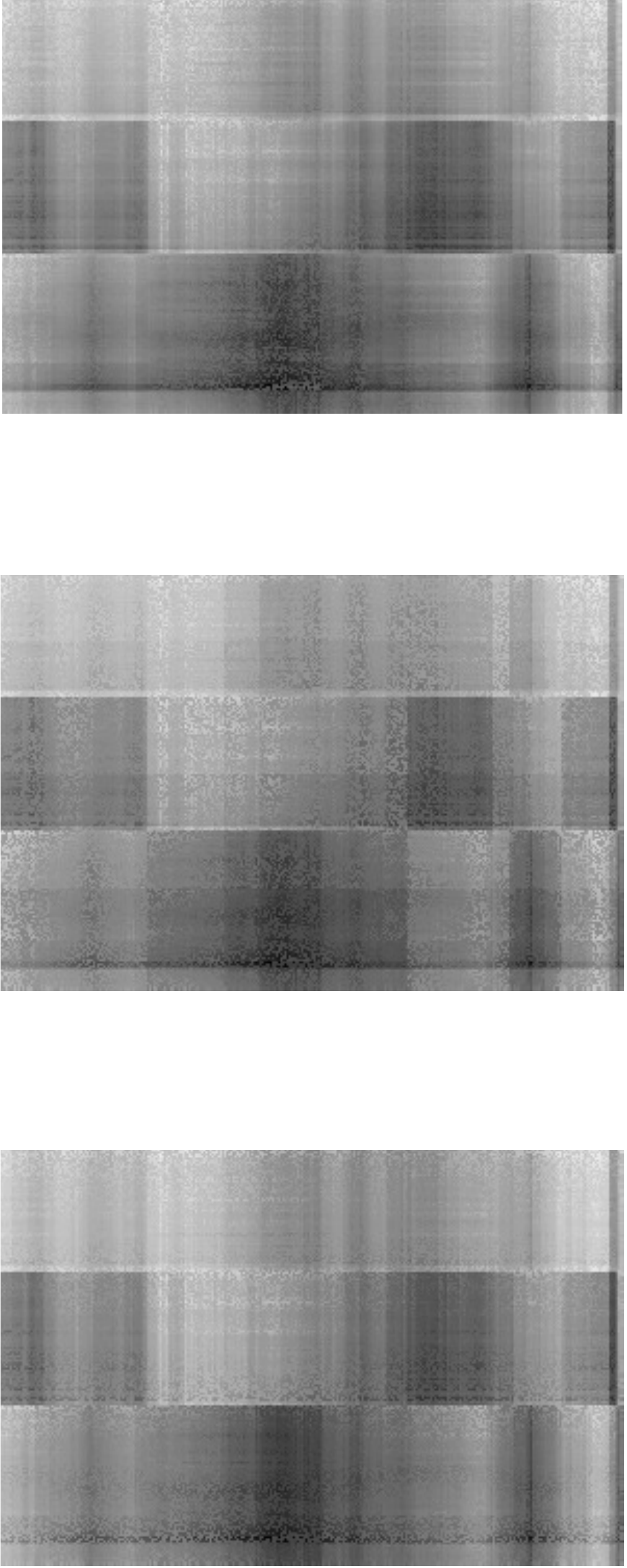}
	\caption{ {\textbf{linkage2}} dataset: Filled-in matrices $\Mtilde{X}$ at multiple scales: $\Mtilde{X}^{(-4,-3)}$,$\Mtilde{X}^{(-1,1)}$,$\Mtilde{X}^{(5,-3)}$ . Rows and columns have been reordered based on the manifold embedding following~\cite{Ankenman2014}.}
	\label{fig:pot2_fill}
\end{figure}

We present the original underlying structure of 3D points used to generate the Euclidean distance matrix $\M{X}$ for the datasets {\textbf{linkage}} and {\textbf{linkage2}} in \Fig{pot} and \Fig{pot2}.
In \Fig{pot_mix} and \Fig{pot2_mix}, on the left we plot the original complete matrix where the rows and columns have been ordered according to the geometry of the 3D points. On the right we plot the matrix we analyze whose rows and columns have been permuted and $50\%$ of the entries have been removed. 
In \Fig{pot_fill} and \Fig{pot2_fill} we display the matrix $\Mtilde{X}^{(l,k)}$ for three pairs of values $l,k$ to demonstrate the smoothing that is occurring across the different scales of the rows and columns.

\end{document}